\documentclass[aps,reprint,superscriptaddress,nofootinbib]{revtex4-2}
\usepackage{main}

\begin{document}

\title{Transient Chaos in BERT}

\author{Katsuma Inoue}
\email{k-inoue@isi.imi.i.u-tokyo.ac.jp}

\affiliation{Graduate School of Information Science and Technology, The University of Tokyo, 7-3-1 Hongo, Bunkyo-ku, Tokyo 113-8656, Japan}

\author{Soh Ohara}
\email{soh.ohara.jp@gmail.com}

\affiliation{Department of Mechanical Engineering, \\
The University of Tokyo, 7-3-1 Hongo, Bunkyo-ku, Tokyo 113-8656, Japan}

\author{Yasuo Kuniyoshi}
\email{kuniyosh@isi.imi.i.u-tokyo.ac.jp}
\affiliation{Graduate School of Information Science and Technology, The University of Tokyo, 7-3-1 Hongo, Bunkyo-ku, Tokyo 113-8656, Japan}
\affiliation{Next Generation Artificial Intelligence Research Center (AI Center), The University of Tokyo, 7-3-1 Hongo, Bunkyo-ku, Tokyo 113-8656, Japan}

\author{Kohei Nakajima}
\email{k\_nakajima@mech.t.u-tokyo.ac.jp}
\affiliation{Graduate School of Information Science and Technology, The University of Tokyo, 7-3-1 Hongo, Bunkyo-ku, Tokyo 113-8656, Japan}
\affiliation{Next Generation Artificial Intelligence Research Center (AI Center), The University of Tokyo, 7-3-1 Hongo, Bunkyo-ku, Tokyo 113-8656, Japan}
\date{\today}

\begin{abstract}

Language is an outcome of our complex and dynamic human-interactions and the technique of natural language processing (NLP) is hence built on human linguistic activities.
Along with Generative Pre-trained Transformer (GPT), Bidirectional Encoder Representations from Transformers (BERT) has recently gained its popularity, owing to its outstanding NLP capability, by establishing the state-of-the-art scores in several NLP benchmarks.
A Lite BERT (ALBERT) is literally characterized as a lightweight version of BERT, in which the number of BERT parameters is reduced by repeatedly applying the same neural network called Transformer's encoder layer.
By pre-training the parameters with a massive amount of natural language data, ALBERT can convert input sentences into versatile high-dimensional vectors potentially capable of solving multiple NLP tasks.
In that sense, ALBERT can be regarded as a well-designed high-dimensional dynamical system whose operator is the Transformer's encoder, and essential structures of human language are thus expected to be encapsulated in its dynamics.
In this study, we investigated the embedded properties of ALBERT to reveal how NLP tasks are effectively solved by exploiting its dynamics.
We thereby aimed to explore the nature of human language from the dynamical expressions of the NLP model.
Our analysis consists of two parts, namely short- and long-term analyses, according to time-scale differences to capture the dynamics.
Our short-term analysis clarified that the pre-trained model stably yields trajectories with higher dimensionality in a certain time range, which would enhance the expressive capacity required for NLP tasks.
Also, our long-term analysis revealed that ALBERT intrinsically shows transient chaos, a typical nonlinear phenomenon showing chaotic dynamics only in its transient, and the pre-trained ALBERT model tends to produce the chaotic trajectory for a significantly longer time period compared to a randomly-initialized one.
Our results imply that local chaoticity would contribute to improving NLP performance, uncovering a novel aspect in the role of chaotic dynamics in human language behaviors.
\end{abstract}

\maketitle
\def\thefootnote{**}\footnotetext{K. I. and S. O. contributed equally to this work}\def\thefootnote{\arabic{footnote}}

\section{Introduction}
Many machine learning models have recently been proposed for natural language processing (NLP) tasks by incorporating deep neural network techniques.
Among them, Bidirectional Encoder Representations from Transformers (BERT) \cite{devlin2018bert}, developed by Google\texttrademark,  have especially gained in popularity, owing to their scalability and efficiency, by establishing the state-of-the-art results for eleven NLP benchmarks, such as the Stanford Question Answering Dataset (SQuAD) 1.1 \cite{rajpurkar2016squad} and general language understanding evaluation (GLUE) benchmark tasks \cite{wang2018glue}.
BERT is a type of feedforward deep neural network architecture composed of multiple encoder layers of Transformers \cite{vaswani2017attention}, by which input sentences are encoded as favorable high-dimensional state vectors potentially capable of solving NLP tasks.
Using self-supervised learning to train network parameters with a large amount of corpus data, BERT is optimized to yield essential representations suitable for multiple NLP tasks.
The \textit{masked language modeling task} (MLM task) is especially used for the \textit{pre-training} process, the objective of which is to predict the original vocabulary ID of the masked word based only on its context.
A huge collection of novel data (BookCorpus) and English Wikipedia articles is used for the pre-training dataset, by which the structure of natural language is expected to be encapsulated in the model.
After the pre-training, an additional simple decoder model, such as logistic regression, is tuned for a specific language task while the pre-trained BERT parameters are fine-tuned or often kept fixed.
This fine-tuning process is completed with a smaller number of training epochs compared to the process of learning from scratch, and thus is computationally inexpensive compared with pre-training.
Despite its simplicity of the training scheme, BERT has outperformed various task-specific architectures at a wide range of NLP tasks \cite{devlin2018bert}.

A Lite BERT (ALBERT) \cite{lan2019albert} is also a powerful neural architecture for NLP tasks, characterized as a lightweight version of the BERT model.
Initially, two types of models, BERT-base and BERT-large, were proposed based on network size, but both are extremely large with more than 100 million parameters.
ALBERT was proposed to reduce BERT parameter size and enhance memory efficiency.
ALBERT incorporates the following two techniques to reduce the number of parameters: (i) factorized embedding parametrization
(the embedding matrix is split into input-level embeddings with relatively low dimensionality, and hidden-layer embeddings with higher dimensionality) and
(ii) cross-layer parameter sharing
(ALBERT shares parameters across layers to improve parameter efficiency).
Even after drastically reducing redundancy, it was reported that ALBERT outperformed the original BERT in typical benchmark tasks \cite{lan2019albert}.

\subsection{ALBERT as ``the reservoir''}
The cross-layer sharing technique provides ALBERT with a structure that reuses the same feedforward neural network.
In other words, ALBERT can be regarded as a type of recurrent neural network (RNN) such that the internal state encodes the generic NLP representation in a specific time step after the input sentence is given.
Also, as with the fine-tuning processes of BERT, additional models (linear models are used in most cases) are trained for each NLP task while the Transformer's encoder parameters are often kept fixed.
This fine-tuning style recalls the \textit{reservoir computing} (RC) framework \cite{maass2002real,jaeger2004harnessing}, in which the internal weights of RNN (a.k.a. \textit{reservoir}) are fixed and only the \textit{readout}'s parameters (often a simple linear regression model) are modified for a specific task \cite{nakajima2020physical,shen2020reservoir} (\cref{fig:1}A).
Moreover, unlike the conventional RC frameworks using random networks, the network parameters are fully tuned to solve the MLM and sentence order prediction task before the readout training phase.
In that sense, the Transformer's encoder can be interpreted as a well-designed reservoir specific for NLP, namely \textit{``the reservoir,''} where the essential properties required in NLP tasks are represented in the form of a dynamical system.
Therefore, we expect that fundamental structures underlying natural language can be anatomized by analyzing the dynamics of the pre-trained ALBERT model.
There have been several works that attempt to analyze the structure of natural language using dynamical systems theory \cite{pollack1991induction,elman1995language,moore1998dynamical} and chaotic dynamics \cite{kataoka2000natural,kramsch2003language,ikegami2003chaotic,mitchener2004chaos}.
Our study aims to advance these dynamical systems approaches to understand natural language, by using the up-to-date machine learning model.

In this study, based on the ``ALBERT as the reservoir'' perspective, we analyze the dynamics of the pre-trained ALBERT and clarify the embedded properties for realizing NLP capability.
In particular, we compare the pre-trained model with a randomly-initialized one and conduct the analysis focusing on two different timescales, namely (i) short-term and (ii) long-term.
In the short-term analysis, we analyze short-term trajectories where the input effects are likely present, by sampling transients within 500 time steps, that is, the outputs of 500 layers.
Here, we show that pre-training ALBERT significantly increases the trajectory's dimensionality while maintaining its coherency, by measuring three indices: synchronization offset, the local Lyapunov exponent (LLE), and effective dimension.
To discuss the contribution of the short-term properties to the representative capability for NLP tasks, we also measure the performances in the short-term range with the following three tasks: MLM task, semantic textual similarity benchmark (STS-B) \cite{cer2017semeval}, and handwriting task (see the Appendix for the detailed setups).
These layerwise analyses are similar to those in \cite{lin2019open} which evaluates BERT performance, and our study inspects the properties for wider time range.
In parallel, we investigate the system's global properties in the long term analysis.
We show that the pre-training increases the chaoticity, that is, the pre-trained trajectory is more likely to magnify small differences in input sentence by measuring the LLE and thereby contributes to amplifying the trajectory's dimensionality.

These properties are fundamentally essential to retain and emphasize differences in input sentences,  i.e., non-chaotic systems cannot properly represent the differences since the information vanishes by converging to a fixed point or limit cycle after a certain time step.
Additionally, we demonstrate that ALBERT intrinsically shows transient chaos \cite{crutchfield1988attractors}, a typical nonlinear phenomenon showing chaotic dynamics only in its transient, and the length of the chaotic trajectory in transient chaos becomes significantly longer due to pre-training.

The significance of this study is clear: we show that pre-training increases the discriminative capability of transients by increasing the chaoticity, which plays an important role in achieving the generic NLP ability of ALBERT by magnifying and retaining the small differences in input sentence, disclosing a novel aspect of information processing in high-dimensional chaotic trajectories.
Finally, we discuss the potential impact of this study and the future direction of studying the dynamical relationship between low-level structures (such as letters, words, phrases, and grammar) and high-level structures (such as logical flow, and meaning).

\subsection{Generalization of ALBERT Model} \label{sec:1.1}
\begin{figure*}[ht]
  \centering
  \includegraphics[width=1.00\textwidth]{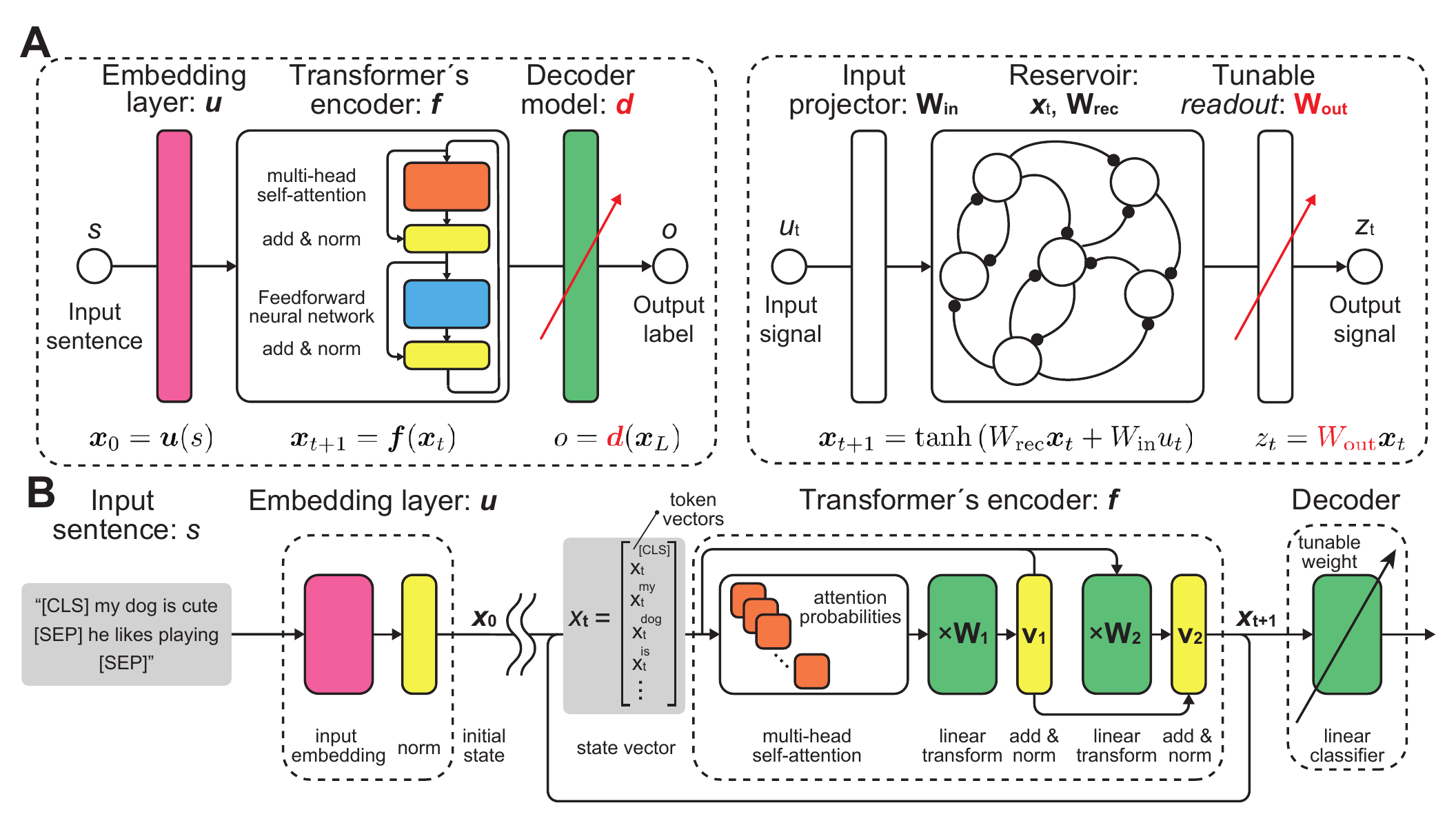}
  \caption[fig1]{
  ALBERT architecture.
  (A) Analogy of the training scheme between the ALBERT architecture and reservoir computing scheme.
  (B) Detailed schematic graph of ALBERT architecture.
  ALBERT can be reinterpreted as a type of discrete-time dynamical system where Transformer's encoder $\bm{f}$ provides the mapping.
  The embedded layer $\bm{u}$ projects input sentence $\bm{s}$ onto the initial state $\bm{x}_0$.
  } \label{fig:1}
\end{figure*}

ALBERT is a deep neural network whose architecture is formulated by the following equations:
\begin{align}
    \bm{x}_{0} &= \bm{u}(\bm{s}) \\
    \bm{x}_{L} &= \underbrace{\bm{f} \circ \bm{f} \cdots \circ \bm{f} }_{L} (\bm{x}_{0})~\left(= \bm{f}^{L}\left(\bm{u}(\bm{s})\right)\right),
\end{align}
where $\bm{u}$ represents the embedding layer and $\bm{f}$ stands for Transformer's encoder (\cref{fig:1}B).
The embedding layer $\bm{u}$ maps input sentence $\bm{s}$ with $N_w$ tokens onto high-dimensional vector $\bm{x}_{0} \in \Real^{N_w\times N_h}$.
Here, $N_h$ is a constant integer called a \textit{hidden dimension} ($N_h=768$ in ALBERT-base, $N_h=1024$ in ALBERT-large) and represents the dimension of token vectors.
The $i$-th token is transformed into the $i$-th row of vector $\bm{x}_{0}^{i}\in\Real^{N_h}~(i=1\cdots N_w)$.
As shown in \cref{fig:1}, Transformer's encoder $\bm{f}:\Real^{N_w\times N_h} \to \Real^{N_w\times N_h}$ also outputs a vector with the same shape as the input vector; that is, $\bm{x}_{L}$ can also be split into $N_w$ vectors $\bm{x}_{L}^{i}$, where $\bm{x}_{L}$ and $\bm{x}_{L}^{i}$ are the \textit{state vector} and \textit{token vector}, respectively.

Transformer's encoder $\bm{f}$ is a feedfoward neural network as well, which includes the attention mechanism \cite{vaswani2017attention} described by the following equation:
\begin{align}
    \bm{f}(\bm{x}) &= v_{2}(\bm{x} W_{2} + v_{1}(A(\bm{x}) W_{1} + \bm{x})),
\end{align}
where $W_1,W_2\in\Real^{N_h \times N_h}$ are weight matrices for linear transformation, $v_{1}, v_{2}$ are the layer normalization \cite{ba2016layer} regulating the values of both the average and variance of elements in each token vector, and $A:\Real^{N_w \times N_h} \to \Real^{N_w\times N_h}$ is a feedforward neural network called \textit{multi-head self-attention}.
In the multi-head self-attention, $N_{a}~(:=N_{h}/64)$ square matrices called \textit{attention probability} ($\in \Real^{N_w\times N_w}$) are calculated, quantifying the strength of relationship among $N_w$ tokens.
The multi-head self-attention $A$ integrates all attention probability matrices to produce an \textit{attention matrix} $A(\bm{x})\in\Real^{N_w\times N_h}$.
During the fine-tuning process, which usually involves logistic regression, an additional model is trained for a specific task based on the output of the $L$-th layer $\bm{x}_{L}$.
In particular, $L=12$ and $L=24$ are used in the ALBERT-base and the ALBERT-large models, respectively, according to the pre-training condition.

This multi-layer feedforward neural network can be described in the form of the following time evolution equation of a discrete dynamical system,
\begin{align}
    \bm{x}_{t+1} &= \bm{f}(\bm{x}_t) \label{equ:ds}.
\end{align}
In other words, the embedding layer $\bm{u}$ can be reinterpreted as a mapping that generates the initial value of the dynamical system from the input sentence, the output of $t$-th layer $\bm{x}_{t}$ as a dynamical state at time $t$.
In this study, we analyzed the dynamics of state vectors and their token vectors to clarify the relationship between the dynamical properties and the NLP capability.

\section{Results}

Based on the formulation of the dynamical system in \cref{sec:1.1}, we investigated the ALBERT dynamics to anatomize the system properties contributing to realizing the generic NLP capability.
As mentioned earlier, to comprehensively understand the mechanism, we analyzed the ALBERT dynamics from the following two different viewpoints; short-term and long-term analysis.
We used a publicly-available pre-trained ALBERT model \cite{lan2019albert}, referred to as the \textit{pre-trained network}, and compared it to a randomly-initialized network, referred to as an \textit{initial network}.
We prepared five initial networks by randomly sampling parameters in the same manner as \cite{lan2019albert} (each weight parameter was independently sampled from a truncated normal distribution $\mathcal{N}(0, \sigma^2)$ where $\sigma=0.02$ was used and the sampled value was clipped in the range of $[-2\sigma, 2\sigma]$. bias parameters are initially set to 0).
Below, statistics for the initial network conditions were calculated using five different networks generated through the above explained procedures.

\subsection{Short-term Analysis} \label{sec:short}
\begin{figure*}
  \centering
  \includegraphics[width=0.90\textwidth]{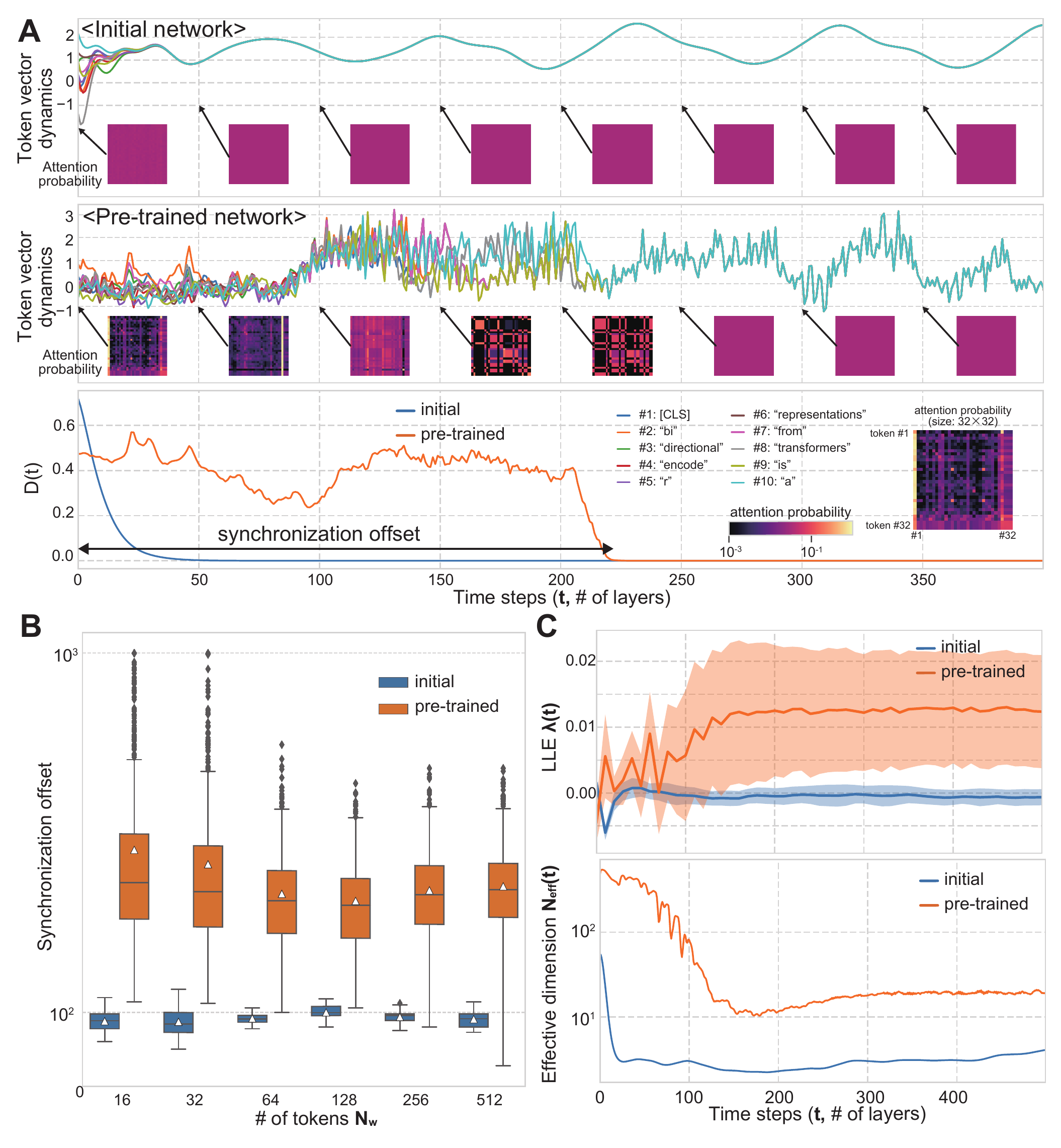}
  \caption[fig2]{
  (A) Token-vector synchronization.
  [Top, Middle] Token vector dynamics.
  We picked up a 32-token sentence from English Wikipedia articles of ``BERT (language model).''
  Only the first element's trajectories of ten token vectors are displayed.
  The corresponding attention probability matrices are also illustrated on the bottom on every 50 time steps.
  [Bottom] Degree of deviation $D(t)$ among the 32 token vectors.
  Synchronization offset is determined by a time step when $D(t)$ becomes lower than a threshold $10^{-5}$.
  (B) Synchronization offsets.
  The triangles represent the average synchronization offsets for the 1,000 input sentences sampled from English Wikipedia articles.
  The outliers outside the upper and lower quartiles are plotted as individual diamond markers.
  (C) Evolution of local Lyapunov exponent $\lambda(t)$ and effective dimension $N_\text{eff}(t)$.
  Both $\lambda(t)$ and $N_\text{eff}(t)$ were measured by 1,000 32-token sentences randomly sampled from English Wikipedia articles.
 } \label{fig:2}
\end{figure*}

\begin{figure*}
  \centering
  \includegraphics[width=0.90\textwidth]{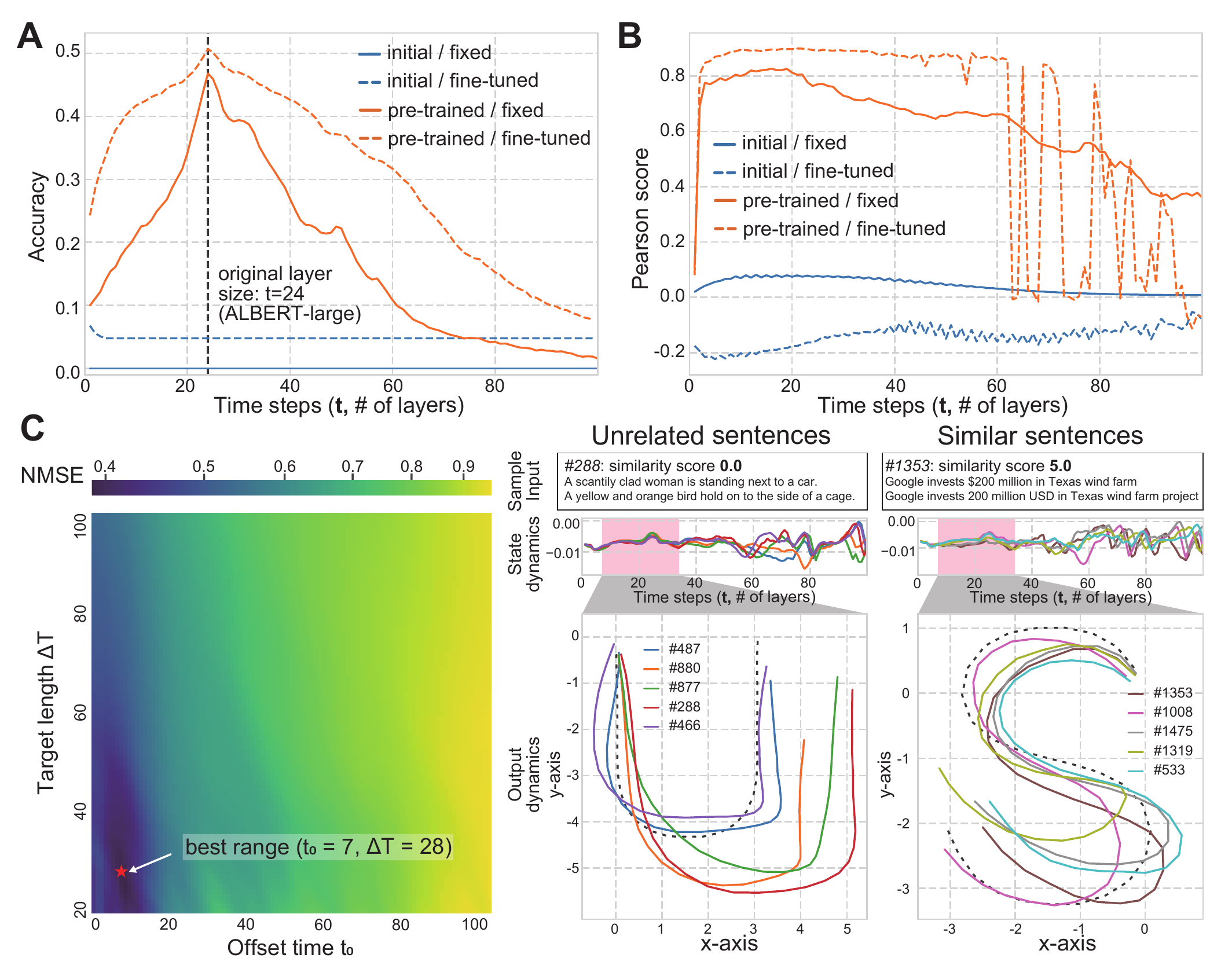}
  \caption[fig3]{
    (A) Performance of MLM task.
    The network parameters were fixed and only the readout parameters were changed.
    We used a publicly-available linear decoder model without fine-tuning (solid lines), while the parameters were tuned in the ``fine-tuned'' conditions (dashed lines).
    (B) Performance of STS-B, one of the GLUE tasks.
    Note that the performance is assessed by the Pearson score between the outputs and targets, which can take negative values.
    In addition to the initial and pre-trained network, we examined the performance of a fine-tuned network where the internal parameters of $\bm{f}$ were tuned (dash lines).
    (C) Handwriting task.
    [Left] Analysis of handwriting task's error.
    Normalized mean square errors (NMSEs) between target and output are displayed in the colormap.
    We changed the offset time $t_0$ and length of target trajectory $\Delta T$, where each linear regression model was tuned by the system's trajectory $\bm{x}_{t}$ ($t \in [t_0, t_0+\Delta T)$).
    [Right] Demonstrations of handwriting task.
    We trained a linear regression model with two output nodes to write the letters ``U'' and ``S'' (black dashed lines) for unrelated and similar sentences, respectively, from the state trajectory $\bm{x}_{t}$ ($t \in [7, 35)$, shaded in pink).
    The corresponding state vector dynamics are also shown in the middle.
    Five best outputs with the smallest errors are displayed in each figure of output dynamics.
  } \label{fig:3}
\end{figure*}

We began by investigating the trajectories in a relatively short-term range within 500 time steps and the contribution to the NLP capability.
Specifically, we first focused on the transients of the high-dimensional token vectors.
\cref{fig:2}A shows the dynamics of the first element of each token vector and the degrees of variation among them.
Using the beginning parts of the article, an English Wikipedia article of ``BERT (language model)'' was selected to prepare the 32-token input sentence.
To evaluate the degree of variation among token vectors, we introduced the following index $D(t)$:
\begin{align}
    D(t) &= \frac{1}{N_w}\sum_{i}^{N_w} ||\bm{x}_t^i - \bar{\bm{x}_t}||^2,
\end{align}
where $\bar{\bm{x}_t}=\frac{1}{N_w}\sum_{i}^{N_w} \bm{x}_t^i$.
Although the more complicated patterns were observed in the pre-trained network compared with the initial one, it was found that all token vectors synchronized ($D(t)\to 0$) after a certain period in both conditions, which we named \textit{token-vector synchronization}.
Token-vector synchronization occurred in all observed cases that we investigated, and all values in the attention probability became identical after token-vector synchronization began.
These observations suggest that the Transformer's encoder and its attention mechanism would have the potential to cause the synchronization phenomenon where all token vectors take the same value.
Note that each token vector element had different convergence values and the timing of synchronization starting, and its value changed according to the input sequence.

As shown in \cref{fig:2}A, the period until the beginning of the token-vector synchronization got significantly longer in the pre-trained network compared with the initial one.
To investigate the contribution of the pre-training to the synchronization timing, we measured the \textit{synchronization offset} with 1,000 sentences randomly sampled from English Wikipedia articles.
Note that synchronization offset was defined as the time step when $D(t)$ falls below threshold $10^{-5}$.
\cref{fig:2}B shows the synchronization offset over token size $N_w$, suggesting that pre-training uniformly extends the synchronization offset regardless of the number of tokens.
This result implies that pre-training amplifies the expressive diversity of the token vector, holding the information about the input sentence for longer time steps.

To further scrutinize and quantify the transient structure, we introduced the following two measures: LLE and \textit{effective dimension} \cite{abbott2011interactions}.
The LLE assesses the expansion degree of the trajectory at each time step $t$.
In particular, the sign of the LLE value works as an indicator discriminating between chaotic and non-chaotic trajectories.
The LLE was calculated in an iterative manner based on the numerical algorithm of the maximum Lyapunov exponent \cite{shimada1979numerical} (see Appendix for the detailed procedures by which to obtain LLE).
The top column of \cref{fig:2}C displays the evolution of averaged values of LLE $\lambda(t)$ under the conditions of $k=1.0, \tau=10$ over 1,000 inputs randomly sampled from English Wikipedia articles, showing that $\lambda(t)$ takes a relatively small positive value in the range of $t<100$ in the pre-training network, while $\lambda(t)$ is uniformly small in the initial one.
This result suggests that the pre-training made the internal dynamics chaotic but embedded a relatively stable structure especially in the short-term range after inputs were given.

The effective dimension evaluates the trajectory's dimensionality, that is, the effective number of the principal components explaining the state cloud (the set of the state vectors).
The effective dimension is calculated from the eigenvalues $\{s_i(t)\}_i~(i=1\cdots N_w \times N_h)$ of co-variance matrix $\mathbb{E}\left[(\bm{x}_t-\bm{\mu}_t)^{T}(\bm{x}_t-\bm{\mu}_t)\right]~(\bm{\mu}_t:=\mathbb{E}[\bm{x}_t])$ with the following equation:
\begin{align}
    N_\text{eff}(t) = \left(\sum_{i} {\tilde{s_i}(t)}^2\right)^{-1},
\end{align}
where $\tilde{s_i}(t)$ represents the normalized eigenvalues so that the sum becomes 1.
In this study, the covariance matrix was calculated from the 1,000 32-token sentences randomly sampled from English Wikipedia articles.
The bigger $N_\text{eff}(t)$ indicates that the state cloud is distributed in the higher-dimensional space and thus the model encodes input sentences in a less redundant manner.
The bottom column of \cref{fig:2}C displays the evolution of $N_\text{eff}(t)$, suggesting that the pre-trained network consistently shows a higher effective dimension than does the initial one.
This steady increase of $N_\text{eff}(t)$ suggests that the input sentences were represented with a wider range of parameters in the state vector as a result of the pre-training, improving separability for input sentences.
A sharp decrease in $N_\text{eff}(t)$ was also observed in the pre-trained network, from $t=50$ to $t=150$, which is consistent with the results in \cref{fig:2}B; that is, the separability is reduced as a result of token-vector synchronization.

So far, we investigated the dynamics in terms of three measures: synchronization offset, LLE, and effective dimension, to analyze the transient structure embedded by pre-training in the short-term range.
It is assumed from the three indices that the pre-training embeds relatively stable transients with higher dimensionality for a longer period compared with the initial one, which would enable the state vector to reflect the difference in input sentences.
(Note that we found that whether this locally stable structure appears or not depends on the type of tasks.
For example, some token-vectors generated from the STS-B dataset yielded chaotic transients with opposite properties to those from the MLM dataset.
See the Appendix for the detailed reports).

In parallel, to illuminate the contribution of these transient properties to the NLP capability, we prepared three NLP tasks and measured the performances.
First, we measured the task performance of the MLM task (see the Appendix for the detailed setups) used in the ALBERT pre-training.
We prepared the publicly-available decoder model of logistic regression tuned in the pre-training procedure.
Originally, the decoder model was optimized to predict masked vocabulary IDs using $\bm{x}_{24}$ for the ALBERT-large model.
Here, we measured the scores of each time step to investigate how the essential information required in the task was encoded and stored in the transients.
We obtained the performance with a \textit{fixed} and \textit{fine-tuned} condition, where the decoder's parameters were unchanged or re-trained for each time step, respectively.
\cref{fig:3}A displays the performances of the MLM task, showing that the pre-training certainly improved the score of the MLM task, at $t=24$ with both the fixed and fine-tuned decoder model.
Notably, the pre-trained accuracy kept relatively high values in a broad range around the original layer size, $t=24$, suggesting that the fundamental structures required to solve the MLM task were constantly held in a certain time range after the pre-training, and not limited in $t=24$.
(Since we used a partial dataset of Wikipedia sentences used in the pre-training, the fine-tuned model slightly outperforms the fixed one, even at $t=24$, by over-fitting to the dataset.
See the Appendix for the detailed setup of the MLM task).

Next, we evaluated the task performance of state vector $\bm{x}_t$ with STS-B.
STS-B is a task in GLUE where the model is tuned to quantify the semantic similarity between two sentences separated by a special token ``[SEP]'' on the scale of 0.0 to 5.0.
The performances of STS-B were measured using the Pearson score with human annotation data, with the performance getting closer to 1.0 with a better model.
In addition to the initial and pre-trained networks with \textit{fixed} parameters of the Transformer's encoder $\bm{f}$, we examined the performance of the \textit{fine-tuned} network where the parameters of $\bm{f}$ were also fine-tuned for the task, verifying the validity of the ALBERT-large model (the learning rates used during the training are shown in the Appendix).
\cref{fig:3}B shows the Pearson score for each $\bm{x}_t$, indicating that the scores with the pre-trained network were constantly higher than those using the initial network, and gradually decreased with larger $t$.
The score did not get significantly higher at $t=24$, implying that the number of layers did not necessarily require designation at the fine-tuning process.
Similarly, in the fine-tuned network, the plateau in score evolution was observed in the range of $t<60$, suggesting performance was independent of the predetermined number of layers.
Note that the performance oscillation in the range of $t>60$ would be caused by the instability of the training for larger architectures, known as the gradient explosion problem \cite{bengio1994learning}, meaning that $t=60$ would be the minimum time step where the gradient values could surpass a preferred range.
These evaluations of STS-B performances imply that the transient of the pre-trained ALBERT model possesses the generic NLP capability in a certain time range and decreases its discriminative capability as time evolves.

The evaluations of the above two tasks show that the pre-trained ALBERT model properly encodes the input sentences into generic state vectors useful to solve NLP tasks in a certain time range.
Finally, we prepared a \textit{handwriting task} to demonstrate that the transient dynamics of ALBERT itself, not the state vector at a certain time step, have a rich expressiveness and can even be exploited as direct computational resource from which to draw required cursive letters.
The handwriting task used the same dataset of STS-B.
Unlike the STS-B task, the decoder model (a.k.a. readout) was tuned to output dynamics of pen directions from the state vector dynamics $\bm{x}_{t}$ in a time range of $t \in [t_0, t_0 + \Delta T)$ to ``write'' the letters ``U'' and ``S''  separately according to the semantic similarities of input sentences (see the Appendix for the detailed setups).
Here, we extracted unrelated and similar sentences whose scores were below 1.0 and over 4.0, respectively, among the original STS-B dataset to create the training and evaluation data.
We first evaluated the expressive performances of the transients by measuring the normalized mean square errors (NMSEs) between the output dynamics and the desired one for each time range.
The colormap in the left part of \cref{fig:3}C indicates that there was a local minimum, and the performance got worse as offset time $t_0$ and target length $\Delta T$ became larger, which is consistent with the result of STS-B performance shown in \cref{fig:3}B, where the performance peaked at $t=16$ and got worse with larger $t$.
The right part of \cref{fig:3}C demonstrates the best five output dynamics for each semantic type of input sentence and the corresponding state vector dynamics of a favorable range scoring minimum NMSE ($t_0=7, \Delta T = 28$).
Note that the same readout was reused to separately ``write'' these letters, suggesting that the high-dimensional transients had sufficient expressiveness to design desired trajectories according to the similarity of input sentences in a certain time range.

To summarize, we analyzed the short-term dynamics of the ALBERT-large model and revealed that pre-training allows the ALBERT architecture to yield locally stable transients with higher dimensionality, especially in the range of $t<100$.
The evaluations of NLP tasks clarified that the transient possesses expressive capabilities reflecting the semantic differences in input sentences in a short-term range, suggesting that the pre-training would enhance the expressive capability for NLP tasks by increasing the transient's dimensionality while keeping its stability moderately.

\subsection{Long-term Analysis} \label{sec:long}
\begin{figure*}
  \centering
  \includegraphics[width=0.90\textwidth]{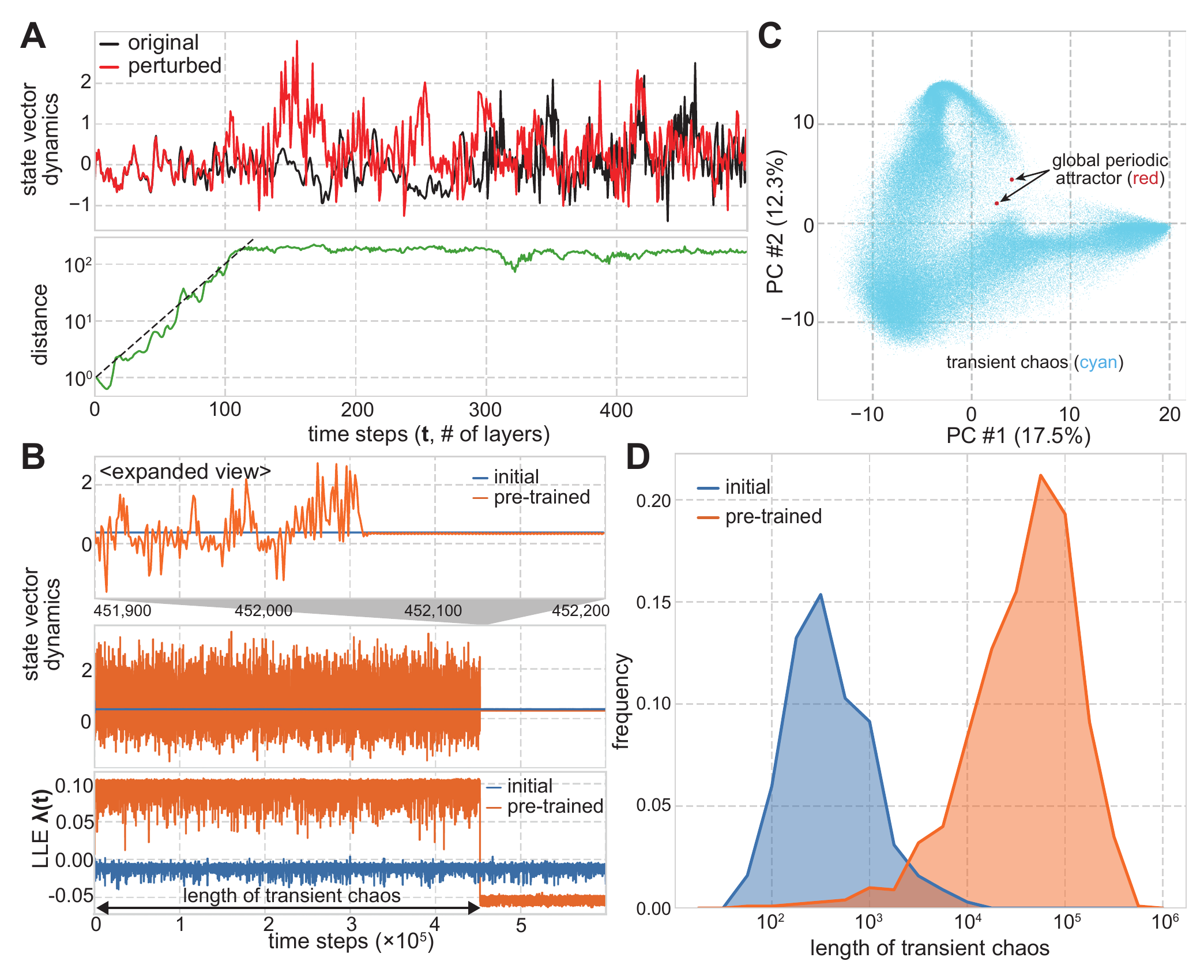}
  \caption[fig4]{
  (A) Effect of perturbation on the initial states.
  [Upper]
  Dynamics of the first elements of state vector with a 32-token input sequence.
  The same sentence used in \cref{fig:2}A was used.
  Both the original dynamics $\bm{x}_t$ (black) and perturbed one $\bm{y}_t$ (red) are shown.
  A small perturbation vector $\bm{\epsilon}~(\|\bm{\epsilon}\|=1.0)$ was added to the original trajectory at $t=0$ (i.e., $\bm{y}_0:=\bm{x}_0+\bm{\epsilon}$).
  [Lower] Evolution of distance $\|\bm{y}_t-\bm{x}_t\|$.
  The slope of the distance expansion (dotted line) corresponds to the value of the LLE.
  (B) Long-term sampling of the trajectory.
  [Top] Expanded view of the first element dynamics for the state vector.
  [Middle] Dynamics of the state vector for 600,000-time steps.
  [Bottom] Corresponding LLE evolution $\lambda(t)$ of the trajectory.
  The value of LLE falls below 0 when the trajectory transits to the periodic orbit or fixed point.
  (C) Principal component analysis of the transient analysis.
  Two primary axes are displayed.
  (D) Distribution of transient chaos length.
  The distribution was obtained by 1,000 32-token sentences randomly sampled from English Wikipedia articles.
  } \label{fig:4}
\end{figure*}

Next, we investigated the long-term trajectories to focus on the global properties of ALBERT-large as a dynamical system.
We especially evaluated the system's chaoticity and investigated pre-training effects on global properties.
First, \cref{fig:4}A demonstrates the original and perturbed trajectories of the pre-training network with the same 32-token sentences used in \cref{fig:2}A.
The exponential expansion of the difference was observed between the original and perturbed trajectory, indicating that the pre-trained ALBERT-large trajectory would be chaotic.

To examine the system's chaoticity more carefully, we sampled extremely long-term trajectories and calculated LLE $\lambda(t)$ again, which offered a measure of local chaoticity at time step $t$ (We used $k=1.0, \tau=50, T=1.0\times10^6$ in this analysis).
\cref{fig:4}B shows dynamics over 600,000 time steps with the same 32-token input sentence, exhibiting the abrupt and unexpected transition from a chaotic trajectory to the periodic orbit.
Notably, the LLE value suddenly fell below 0 after the transition, indicating that the chaotic trajectory transited to a non-chaotic one.
This characteristic phenomenon showing chaotic dynamics only in its transient is known as \textit{transient chaos} \cite{crutchfield1988attractors} and is often observed in high-dimensional dynamical systems.
In addition, all dynamics, as far as we observed, transited to the same periodic global attractor, implying that the pre-trained network was globally non-chaotic and possessed a single global periodic attractor basin.
We also analyzed the obtained trajectory by principal component analysis in \cref{fig:4}C, suggesting the trajectory of transient chaos had a certain structure and was distributed on a high-dimensional state space.

To evaluate the effect of pre-training on the transient chaos length, we measured the distribution of the transient chaos length using 1,000 32-token sentences randomly sampled from English Wikipedia articles.
\cref{fig:4}D shows the distribution of the transient chaos length, indicating that the system was more likely to produce chaotic trajectories for a longer period after pre-training.
The initial network failed to produce chaotic trajectories with 38.5\% of tested sentences, while all samples yielded transient chaos in the pre-trained network.
The average LLE value in chaotic trajectories was close to zero in the initial network ($6.41\times 10^{-3}$), yet positive in the pre-trained network ($9.54\times 10^{-2}$).
These results suggest that pre-training increases local chaoticity and extends the length of chaotic trajectories.

\section{Discussion}

In this study, based on the formulation of ALBERT as a discrete-time dynamical system, we analyzed the trajectory properties using short-term and long-term analyses.
In the short-term analysis, we first demonstrated that token vectors began to synchronize after a certain period, and suggested that it was caused by the attention mechanism of the Transformer's encoder.
Recent studies have showed that graph neural networks \cite{zhou2018graph}, including the ALBERT's attention mechanism, can cause \textit{over-smoothing} and eliminate the difference among nodes with the deep architecture \cite{li2018deeper,oono2019graph}, which is consistent with our empirical results exhibiting token-vector synchronization.
We also measured three indices to quantify the transient properties and found that:
\begin{itemize}
    \item pre-training significantly extended the synchronization offset,
    \item local chaoticity was low, while the effective dimension was significantly high, especially when $t$ was small for sentences used in pre-training (English Wikipedia sentences).
\end{itemize}
These results suggest that the pre-trained ALBERT generates a relatively stable trajectory, with higher dimensionality in a certain short-term range.
Moreover, the gradual changes in the benchmark scores around $L=24$ indicate that NLP functionality is not implemented by the designated composite function $\bm{f}^{L}$; instead, it is formed by a single mapping $\bm{f}$, and essential structures required in NLP tasks are maintained in its transients for a certain period.
This property allows the transients to design the output dynamics according to the meaning of input sentences, which was demonstrated in the handwriting task.
Ref. \cite{liu2019linguistic} also reported that optimal performance was not always obtained in the predetermined number of layers (e.g., $t=24$ for ALBERT-large model), which is also compatible with our results.
Since the NLP performances were especially high when using short-term transients of $t<50$, it is assumed that the discrimination ability of ALBERT for NLP tasks is enhanced by diverse trajectory patterns induced by pre-training.

Next, the long-term analysis showed that transient chaos was generated for a longer period through pre-training.
It was reported that nonlinear dynamical systems become universally unstable over a certain dimensionality \cite{ishihara2005magic}.
In ALBERT architecture, the nonlinearity of the system is provided by a softmax function in the multi-head self-attention layer, layer normalization, and GLUE activation.
Therefore, it is quite possible that chaotic trajectories emerge on ALBERT architecture.
In addition, LLE and transient chaos length values revealed that pre-training not only increases the local chaoticity but also produces a chaotic trajectory for a longer period.
Since the effective dimension generally increases in the chaotic regime \cite{abbott2011interactions}, it was considered that local chaoticity induced by pre-training contributes to increasing the trajectory's dimensionality, enabling the system to amplify input sentence differences and consequently enhance separability for NLP tasks.

Furthermore, the coexistence of long-term chaotic dynamics and locally stable trajectories would be a significant property when utilizing chaos in information processing.
For example, an RNN framework called \textit{innate training} \cite{laje2013robust} revealed a novel aspect of a chaotic trajectory for information processing.
Similar to ALBERT's self-supervised algorithm, the internal weights of the chaotic RNN are adjusted by a semi-supervised learning scheme to reproduce the chaotic dynamics once generated by itself.
As a result, the trained RNN can reproduce versatile spatiotemporal chaotic dynamics according to the input, which can be used for forming and controlling various transient patterns \cite{laje2013robust,goudar2018encoding,inoue2020designing}.
In that sense, it can be said that ALBERT exploits the chaotic trajectory in information processing with a mechanism analogous to innate training.

We formulated ALBERT as a dynamical system and investigated dynamical properties.
This generalization greatly expands applications of ALBERT's architecture.
For example, ALBERT can be applied or expanded to time series processing tasks.
As demonstrated in the handwriting task, motor command generation and natural language recognition can be naturally implemented in a single dynamical system, exhibiting the possible use of ALBERT at the real-world environment.
Also, the existence of chaoticity in ALBERT offers a new guideline for the application of nonlinear chaotic dynamical systems.
Conventionally, researchers have mainly focused on studying ways to suppress and control the system's chaoticity \cite{ott1990controlling,pyragas1992continuous,shinbrot1993using}.
Our results may imply the positive contributions of chaotic dynamics to human linguistic activities by shedding light on the application of chaotic dynamics to information processing.

To investigate the general properties of ALBERT, we randomly chose sentences from English Wikipedia articles without any rules.
We also did not directly change the content of the input sequence.
However, ALBERT can properly recognize the difference in the meanings of input sentences as shown via several benchmark tasks.
In other words, it is speculated that ALBERT can adjust the degree of chaoticity according to the sentence meaning.
The discriminating capability of sentence meaning raises an important question: which of ALBERT's properties enable the system to reflect the  difference in sentence meanings in the dynamics?
For example, in our daily lives, we can distinguish between sentences having opposite meanings with a single word flip, which is a chaotic-like state sensitivity.
Conversely, we can comprehend that completely different texts, the ordering of words, can generate the same meaning.
Here, the relationship between the sentences and their meanings should be nonlinear.
We expect that ALBERT might express this nonlinear correspondence by exploiting embedded chaoticity.
If this is the case, we can quantitatively evaluate the relationship between sentence meaning and dynamical properties by directly manipulating the input sentences for ALBERT, which yields significant implications for the relationship between symbolic structure and sentence meaning in human language processing.
These points are explored in our future works.

\section*{Acknowledgement}
This work was based on results obtained from a project commissioned by the New Energy and Industrial Technology Development Organization (NEDO).
K. I. was supported by JSPS KAKENHI Grant Number JP20J12815.
K. N. was supported by JSPS KAKENHI Grant Number JP18H05472, and by JST CREST Grant Number JPMJCR2014, Japan.

\section*{Appendix}
\renewcommand{\thefigure}{A\arabic{figure}}
\renewcommand{\thealgorithm}{A\arabic{algorithm}}
\setcounter{figure}{0}
\begin{figure*}
  \centering
  \includegraphics[width=0.90\textwidth]{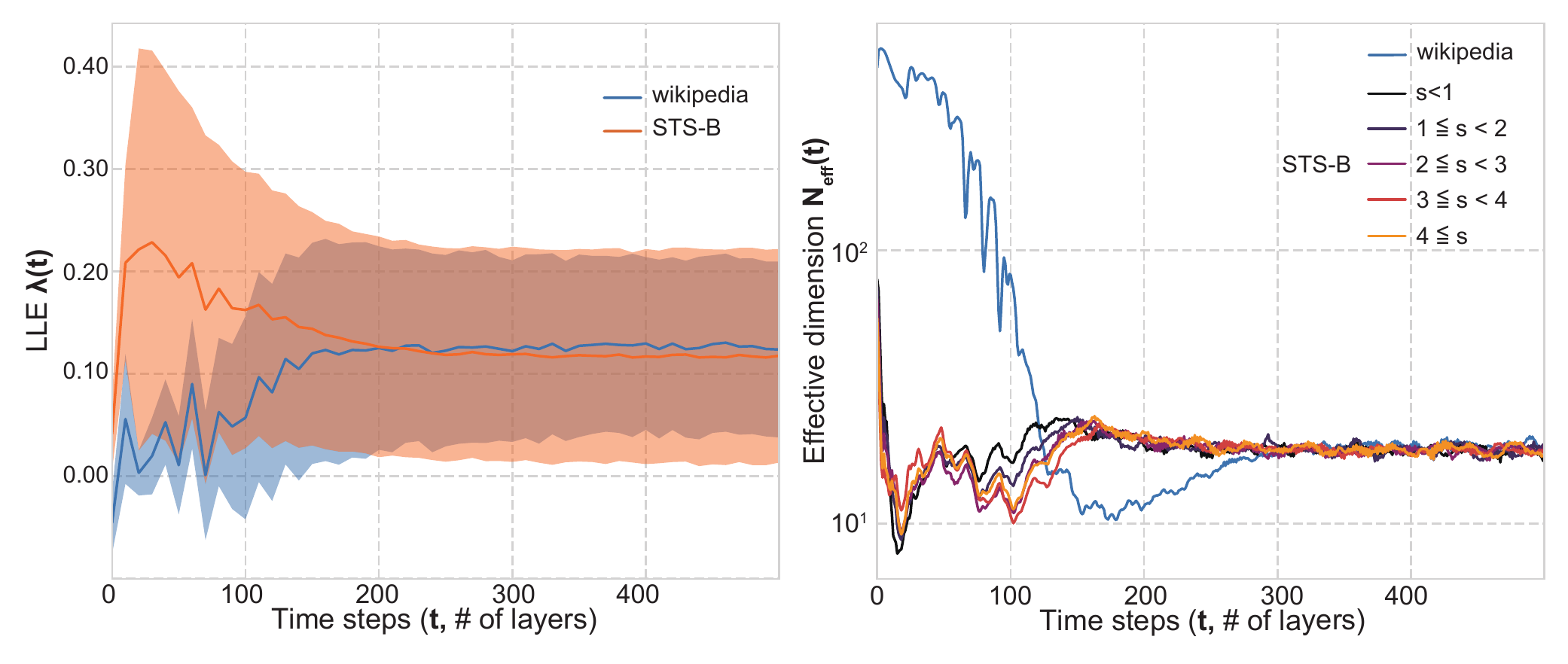}
  \caption[fig:a1]{
  Trajectory analysis for the STS-B dataset.
  [Left] Comparison of LLE between STS-B sentences and Wikipedia sentences used in MLM task.
  [Right] Evolution of effective dimension.
  To investigate the effect of sentence meaning on the dynamics, we divided STS-B sentences into 5 classes according to the annotated semantic similarity score.
  } \label{fig:a1}
\end{figure*}
\subsection{Dataset used in MLM task}
MLM is a pre-training task of BERT and ALBERT.
The network parameters are trained to predict the original vocabulary ID of the masked word based only on its context.
Originally, 58,476,370 sentences were prepared for pre-training by extracting the beginning parts of the English Wikipedia articles.
In this study, we extracted 10,000 sentences from the dataset and used 9,500 sentences for the training and 500 sentences for the evaluation.

\subsection{Detailed setups of the STS-B training}
STS-B is one of the GLUE tasks used to evaluate the performance for quantifying the semantic similarity between two sentences separated by a special token ``[SEP]'' on the scale of 0.0 to 5.0.
For obtaining the results displayed in \cref{fig:3}C, we set the learning rates to $10^{-3}$ and $10^{-6}$ for the fixed and fine-tuned setups, respectively.
We tuned the parameters for 10 epochs.

\subsection{Algorithm for local Lyapunov exponent}
Local Lyapunov exponent (LLE) $\lambda(t)$ for a dynamics $\bm{x}_t$ with an initial value $\bm{x}_0$ is calculated with an iterative algorithm (\cref{algorithm:lle}).

\subsection{Trajectory analysis for STS-B dataset}
To understand the relationship between sentence meaning and the dynamics properties, we also measured the LLE and effective dimensions for the STS-B dataset.
The analysis of the LLE shown in the left part of \cref{fig:a1} shows that the STS-B's transients are more likely to become chaotic in the short-term range.
Also, evaluations of the effective dimension displayed in the right part of \cref{fig:a1} clarify that the dimensionality of the transient becomes lower in the short-term range.
These properties by the STS-B dataset are opposites to those by the Wikipedia one, and indicate that the locally stable structure obtained by the pre-trained dataset was not necessarily observed, and this outcome depends on the type of tasks and corresponding input structures.

\subsection{Detailed setups of the handwriting task}
The handwriting task evaluates the expressive capability of a transient in a certain time range, where the external linear regression model is tuned to separately write the letters ``U'' and ``S'' according to the semantic similarity of sentences from the STS-B dataset.
The base paths ``U'' and ``S'' were prepared from Arial font.
The target dynamics of the pen directions were calculated by interpolating the paths with $\Delta T + 1$ points.
The readout models were tuned to output the dynamics of the pen direction $\bm{v}_t = [\Delta x_t, \Delta y_t]^T\in \mathcal{R}^{2}$ and the accumulated trajectories $\bm{p}_t = \sum_{k=t_0}^{t} \bm{v}_{t}$ are shown in \cref{fig:3}C.
The normalized mean square errors (NMSEs) between the target $\bm{d}(t)$ and the output $\bm{y}(t)$ were calculated as follows:
\begin{align*}
    \text{NMSE}(\bm{y}, \bm{d}) = \left< \frac{\sum_{t \in [t_0, t_0+\Delta T)} \|\bm{y}(t)-\bm{d}(t)\|^2}{\sum_{t \in [t_0, t_0+\Delta T)} \|\bm{d}(t)\|^2} \right>
\end{align*}
where the bracket represents the average over the evaluation data.

\begin{algorithm}[H]
  \caption{Local Lyapunov Exponent (LLE)}
  \begin{algorithmic}[1]
    \State $\bm{x}_0 = \bm{u}(\bm{s})$ \Comment{Calculating $\bm{x}_0$ for input sentence $s$}
    \State $\bm{\epsilon} \sim \mathcal{N}(\bm{0}, I)$ \Comment{Sampling perturbation $\bm{\epsilon}$}
    \State $t \leftarrow 0$ \Comment{reset time $t$}
    \While{$t < T$}
        \State $\bm{\epsilon} \leftarrow k\bm{\epsilon}/\|\bm{\epsilon}\|$ \Comment{Normalizing $\bm{\epsilon}$}
        \State $\bm{y}_t = \bm{x}_t + \bm{\epsilon}$
        \State $\lambda(t) = \ln \frac{\|\bm{y}_{t+\tau} - \bm{x}_{t+\tau}\|}{\|\epsilon\|}$ \Comment{LLE for time $t$}
        \State $\bm{\epsilon} \leftarrow \bm{y}_{t+\tau} - \bm{x}_{t+\tau}$ \Comment{Recalculating perturbation $\bm{\epsilon}$}
        \State $t \leftarrow t + \tau$
    \EndWhile
  \end{algorithmic} \label{algorithm:lle}
\end{algorithm}

\nocite{*}
\bibliographystyle{bib/IEEEtransBST/IEEEtran}
\bibliography{main}

\end{document}